# Improvement of K Mean Clustering Algorithm Based on Density

Su Chang,[1] Xu Zhenzhong,[2] Gao Xuan,[3]

[1]*School of Foreign Languages, Harbin Institute of Technology, Harbin 150001, PR China*
[2]*College of Computer Science and Technology, Beihang University, Beijing 100191, PR China*
[3]*State Key Laboratory of Superhard Materials, Jilin University, Changchun 130000, PR China*

**Abstract**

The purpose of this paper is to improve the traditional K-means algorithm. In the traditional K mean clustering algorithm, the initial clustering centers are generated randomly in the data set. It is easy to fall into the local minimum solution when the initial cluster centers are randomly generated. The initial clustering center selected by K-means clustering algorithm which based on density is more representative. The experimental results show that the improved K clustering algorithm can eliminate the dependence on the initial cluster, and the accuracy of clustering is improved.

**Introduction**

Clustering is an important research field in data mining and it is a scientific and effective algorithm for studying clusters and obtaining useful information. Clustering algorithm is a kind of unsupervised learning, which is widely used. It has broad application prospects in image processing, pattern recognition, market analysis, target customer orientation, biological population division and so on.

The K-means algorithm proposed by J.B.MacQueen in 1967 is the most widely used and the most mature clustering algorithm [1]. K-means clustering algorithm is a classical classification clustering algorithm, and it is also an iterative clustering algorithm. In the course of iteration, the cluster center is moved continuously until the clustering criterion function converges.

The basic idea of K-means clustering algorithm in the data set: select k data objects randomly as k initial cluster centers. According to the mean object in the class, namely the cluster center, divide other data objects in turn to the nearest cluster center with the class. After completing the division of the data image, calculate each cluster, update the clustering center as the new clustering center. The above clustering process is iterated until the cluster center does not change any more, i.e.. The convergence of the clustering criterion function value is so high that the criterion function E is defined as:

$$E = \sum_{i=1}^{k} \sum_{x \in C_i} |x - \overline{x_i}|^2 \qquad (1)$$

The larger the E is, the greater the distance between the object and the cluster center, the lower the similarity in the cluster ; on the contrary, the lower the E is, the higher the similarity within the cluster . $X_i$ is the clustering center of $C_i$; K is the number of clusters; $C_i$ is the i[th] cluster in K clusters.

The main drawback of K-means algorithm is the great dependence of the initial clustering center. Due to the random selection of initial cluster center, the algorithm often falls into local minima. (1) is a Non-convex function, there are often many local minima in Non-convex functions, but only one of them is the global minimum. In this paper, consider density and distance based on density K-means clustering algorithm. On the one hand, choosing larger density points as the clustering center can reduce the interference of noise points and reduce the number of iterations to improve the operation efficiency. On the other hand, it can avoid choosing too many data points in the same cluster when selecting the initial cluster center, while other clusters (mainly small clusters) have no cluster centers. That is to say, the initial cluster center selection is too close, resulting in bad clustering results.

**Improved K mean algorithm based on density algorithm**

The traditional K mean algorithm is sensitive to the initial cluster center, and different initial cluster centers often have a greatly impact on the final clustering results, which may lead to high volatility. In view of this shortcoming, we must select the data which can reflect the data distribution characteristics as the initial clustering center to optimize the algorithm, and improve the stability of the algorithm.

In the traditional K-means algorithm, people need to specify K value in advance, which will lead to the final clustering results to a certain extent is not accurate. The proposed algorithm based on density to improve k-means will objectively reflect the types of clustering results according to the data features. The core of the improved clustering algorithm has the following two characteristics:

1. It's surrounded by the point of which density is less than it.
2. The distance between the core and other larger data points is larger.

Considering the data set to be clustered as the corresponding index set to express the distance between every data point. For any data point in S, two quantities can be defined, and these two quantities are used to characterize the two characteristics of the clustering center mentioned above. Two methods of calculating local density are given below:

Local density $\rho_i$:

Cut-off kernel

$$\rho_i = \sum_{j \in I_{s \setminus \{i\}}} \chi(d_{ij} - d_c) \quad (2)$$

Function

$$\chi(x) = \begin{cases} 1, & x < 0 \\ 0, & x \geq 0 \end{cases} \quad (3)$$

The parameter $d_c > 0$ is the truncation distance, which needs to be specified beforehand. According to the empirical value, a $d_c$ is selected, so that the number of points around each data point is about 1%~2% of the total number of data points. For each point in the data, there are distances between the data points and the other N-1 data points. A total of N (N-1) distances are found, half of which are repeated. The distance *dij (i <j)*, a total of *M = 1 / 2N (N-1)* points, sorted from small to large to get the resulting sequence $d_1 \leq d_2 \leq \cdots \leq d_m$. Taking $d_c$ as $d_k$, $k \in \{1, 2, ..., M\}$. From the overall view of N (N-1) distance, the distance that satisfies the "less than $d_c$" condition is about k/M, that is about (k/M) N (N-1). The average to each data point is (k/M) N. The proportion here is equivalent to the t in the algorithm.

$$d_c = d_{f(Mt)}$$

Where $f_{(Mt)}$ is an integer after rounding the Mt. According to (2) we can see that the number of data points between S and $x_i$ is less than $d_c$, and the number of $x_i$ itself is not considered here.

Gaussian kernel

$$\rho_i = \sum_{j \in I_{s \setminus [i]}} e^{-(\frac{d_{ij}}{d_c})^2} \tag{4}$$

It can be found by comparison that cut-off kernel is discrete, and Gaussian kernel is continuous. In comparison, the formula (4) means that the probability of the same local density of different data points is smaller. For equation (4), there still exist "more data points less than $d_c$ from $x_i$, the greater the value of $\rho_i$

Distance $\delta_i$:

Let x indicate a descending order of subscript, it satisfies

$$\rho_{q1} \geq \rho_{q2} \geq \cdots \geq \rho_{qN} \tag{5}$$

It can be defined:

$$\delta_{qi} = \begin{cases} \min_{\substack{qj \\ j<i}} \{d_{qiqj}\}, & i \geq 2; \\ \max_{j \geq 2} \{\delta_{qj}\}, & i = 1. \end{cases} \tag{6}$$

When $x_i$ has the largest local density, it represents the maximum distance between the data point in S and $x_i$; otherwise, it represents the minimum distance between $x_i$ and that (or those) data points whose local density is greater than $x_i$.

So far, for each data point x in S, we can calculate $(\rho_i, \delta_i), i \in I_s$. Considering the example in Figure 1 (A), which contains 20 two-dimensional data points, and draw the decision diagram by drawing two pairs which considers $\rho$ as the horizontal axis and $\delta$ as the vertical axis.

As shown in Figure 1 (B)

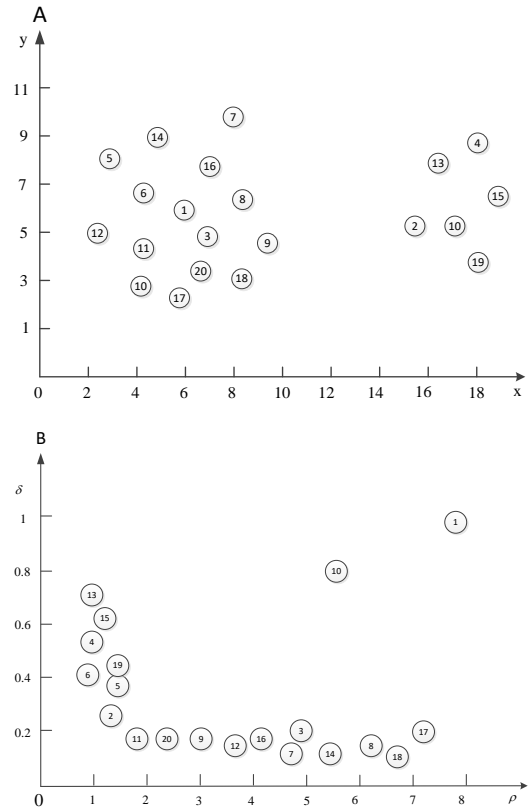

Fig. 1 example and schematic diagram

It can be seen from the graph that the data points of first and tenth have larger values and values, and the two data points correspond to the clustering centers shown in the graph (1A). At the same time, it can be seen that the three data points numbered 4, 13 and 15 are outliers in the original data set. They are characteristic in Figure 1 (B), δ is very large, but ρ is very small.

Graph 1 (B) has a decisive effect on the selection of clustering centers, that is, qualitative analysis is carried out when selecting the cluster centers, which contains large subjective factors, and different cluster centers may be obtained according to the same graph. Now we will calculate the value and get a comprehensive consideration.

$$\gamma_i = \rho_i \delta_i, \quad i \epsilon I_s \qquad (7)$$

It can be seen from formula 7 obviously that the bigger the $\gamma_i$ value is, the more likely it is the clustering center. It will be arranged in descending order, and then several data points will be truncated as clustering centers.

For the number of cluster centers, the calculated values are plotted according to the above values. As shown in Figure 2

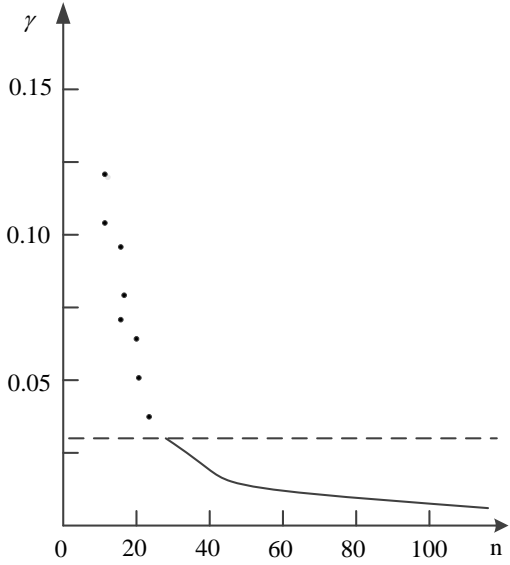

Figure 2. A schematic diagram of results in descending order

It can be seen from the figure that the value of the non-clustering center is more smooth, and there is a significant jump from the non-clustering center to the cluster center. According to the above, the clustering center can be easily obtained.

In this paper, the improved K-means algorithm will use the kernel function distance Euclidean distance, thus solving the shortcomings, that K mean clustering algorithm is difficult to find other clusters of spherical clusters. The kernel function used in this paper is Conditional positive definite kernel. If a symmetric function K: K: $X \times X \to R, n \epsilon N, x_i \epsilon X$ and $c_i \epsilon R$, when satisfied $\sum_{i=1}^{n} c_i = 0$:

$$\sum_{i,j=1}^{n} c_i c_j k_{ij} \geq 0 \qquad (8)$$

K is called conditionally positive definite kernel. Therefore, the conditional positive definite kernel only needs to satisfy and formula (8), and can be calculated in kernel operations.

Inner product is the most basic element of vector similarity, so the inner product function can be considered as a simplified measure of similarity, which is used in classification recognition.

Positive definite kernel function [9]:

$$k(x_i, x_j) = -||x_i - x_j||^q \qquad (9)$$

When $\sum_{i}^{n} c_i = 0$, there is:

$$-\sum_{i,j=1} c_i c_j ||x_i - x_j||^q$$
$$= -\sum_i c_i \sum_j c_j ||x_j||^q - \sum_j c_j \sum_i c_i ||x_i||^q + 2\sum_{i,j} c_i c_j <x_i, x_j>$$
$$= 2\sum_{i,j} c_i c_j <x_i, x_j>$$
$$= 2||\sum_i c_i x_i||^q \geq 0$$

当 $0 \leq q \leq 2$ 时，$k(x_i, x_j)$

At that time, it can be used in nonlinear classification algorithm.

The improved algorithm is applied to the K mean algorithm, and the calculation steps based on the density improved K-means algorithm are as follows:

Input: sample data

1) Choosing the clustering center based on the density, and selecting the number of clusters according to the decision diagram.

2) The kernel function is used to calculate the distance between the cluster center and other data points, and the distance is used to determine which cluster center the specific data points belong to.

3) Calculating the criterion function $E = \sum_{i=1}^{k} \sum_{x \in C_i} |x - \bar{x}_i|^2$, that is, the value of the sum of squares of error. Compare the accuracy of the two algorithms.

**Experimental results and analysis**

The experimental data used in this paper is the data in the UCI database. Hayes-roth, Wine, Iris data are used in the database. UCI database is a common database [10] for specialized machine learning and data mining algorithms. The data in the library has a definite classification, so it can express the quality of clustering intuitively. In order to verify the accuracy of the algorithm, the data distribution of the test data set is kept in the original state without any manual processing.

The Iris, Wine, and Hayes-roth raw data are listed as follows:

Table 1 Iris data

| Iris | | | | |
|---|---|---|---|---|
| sepal length | sepal width | petal length | petal width | class |
| 5.1 | 3.5 | 1.4 | 0.2 | Iris-setosa |
| 4.9 | 3.0 | 1.4 | 0.2 | Iris-set：osa |
| 4.7 | 3.2 | 1.3 | 0.2 | Iris-setosa |
| 4.6 | 3.1 | 1.5 | 0.2 | Iris-setosa |
| 5.0 | 3.6 | 1.4 | 0.2 | Iris-setosa |
| 5.4 | 3.9 | 1.7 | 0.4 | Iris-setosa |
| 7.0 | 3.2 | 4.7 | 1.4 | Iris-versicolo |
| 6.4 | 3.2 | 4.5 | 1.5 | Iris-versicolo |
| 6.3 | 3.3 | 6.0 | 2.5 | Iris-virginica |
| 5.8 | 2.7 | 5.1 | 1.9 | Iris-virginica |

Table 2　Wine data

| Wine | | | | | | | | | | |
|---|---|---|---|---|---|---|---|---|---|---|
| Ash | Alcalinity of ash | Magnesium | Total phenols | Flavanoids | Nonflavanoid phenols | Proanthocyanins | Color intensity | Hue | OD280/OD315 of diluted wines | Proline |
| 1.71 | 2.43 | 15.6 | 127 | 2.8 | 3.06 | 0.28 | 2.29 | 5.64 | 1.04 | 1065 |
| 2.14 | 11.2 | 100 | 2.65 | 2.76 | 0.26 | 1.28 | 4.38 | 1.05 | 3.4 | 1050 |
| 2.67 | 18.6 | 101 | 2.8 | 3.24 | 0.3 | 2.81 | 5.68 | 1.03 | 3.17 | 1185 |
| 2.5 | 16.8 | 113 | 3.85 | 3.49 | 0.24 | 2.18 | 7.8 | 0.86 | 3.45 | 1480 |
| 2.87 | 21 | 118 | 2.8 | 2.69 | 0.39 | 1.82 | 4.32 | 1.04 | 2.93 | 735 |
| 2.4 | 23 | 101 | 2.83 | 2.55 | 0.43 | 1.95 | 2.57 | 1.19 | 3.13 | 463 |
| 2 | 19 | 86 | 2.2 | 2.53 | 0.26 | 1.77 | 3.9 | 1.16 | 3.14 | 714 |
| 2.2 | 18.8 | 86 | 2.2 | 0.3 | 1.43 | 2.5 | 0.26 | 1.77 | 3.9 | 630 |
| 2.51 | 24 | 78 | 2 | 1.58 | 0.4 | 1.4 | 2.2 | 2.53 | 3.05 | 520 |
| 2.32 | 22.5 | 85 | 1.65 | 1.59 | 0.61 | 1.62 | 4.8 | 0.84 | 2.01 | 515 |
| 2.48 | 23 | 102 | 1.8 | 0.75 | 0.43 | 1.41 | 7.3 | 0.7 | 1.56 | 750 |
| 2.26 | 20 | 120 | 1.59 | 0.69 | 0.43 | 1.41 | 7.3 | 0.7 | 1.56 | 750 |
| 2.37 | 20 | 120 | 1.65 | 0.68 | 0.53 | 1.46 | 9.3 | 0.6 | 1.62 | 840 |
| 2.74 | 24.5 | 96 | 2.05 | 0.76 | 0.56 | 1.35 | 9.2 | 0.61 | 1.6 | 560 |

Table 3  Hayes-Roth data

| Hayes-Roth | | | | | |
|---|---|---|---|---|---|
| name | hobby | age | educational | matrial | class |
| 92 | 2 | 1 | 1 | 2 | 1 |
| 10 | 2 | 1 | 3 | 2 | 2 |
| 83 | 3 | 1 | 4 | 1 | 3 |
| 61 | 2 | 4 | 2 | 2 | 3 |
| 107 | 1 | 1 | 3 | 4 | 3 |
| 113 | 1 | 1 | 3 | 4 | 3 |
| 80 | 3 | 1 | 3 | 2 | 2 |
| 125 | 3 | 4 | 2 | 4 | 3 |
| 36 | 2 | 2 | 1 | 1 | 1 |
| 105 | 3 | 2 | 1 | 1 | 1 |
| 81 | 1 | 2 | 1 | 1 | 1 |
| 122 | 2 | 2 | 3 | 4 | 3 |
| 94 | 1 | 1 | 2 | 1 | 1 |
| 60 | 2 | 1 | 2 | 2 | 2 |
| 8 | 2 | 4 | 1 | 4 | 3 |
| 20 | 1 | 1 | 3 | 3 | 1 |
| 85 | 3 | 2 | 1 | 2 | 2 |
| 50 | 1 | 2 | 1 | 1 | 1 |

In the improved K-means clustering algorithm, the initial cluster center is determined, so the experiment only works once after the improvement, the highest, minimum and average clustering accuracy as well as the maximum, minimum and average criterion function E are the same. In this paper, the traditional frontal K mean algorithm is tested 20 times.

Table 4 Comparison of clustering accuracy

| Algorithm | Database | Clustering accuracy | | |
|---|---|---|---|---|
| | | maximum | minimum | average |
| k-means | Iris | 89.43 | 82.34 | 85.89 |
| | Wine | 78.56 | 75.67 | 77.12 |
| | Hayes-Roth | 81.23 | 70.43 | 75.83 |
| Algorithm in this paper | Iris | 90.23 | 90.23 | 90.23 |
| | Wine | 75.78 | 75.78 | 75.78 |
| | Hayes-Roth | 82.45 | 82.45 | 82.45 |

Table 5 Comparison of criterion functions

| Algorithm | Data | $E_{max}$ | $E_{min}$ | $E_{avg}$ |
|---|---|---|---|---|
| k-means | Iris | 145.28 | 78.94 | 112.11 |
|  | Wine | 2.7878 | 2.34 | 2.56 |
|  | Hayes-Roth | 2.14 | 1.89 | 2.022 |
| Algorithm in this paper | Iris | 78.94 | 78.94 | 78.94 |
|  | Wine | 2.34 | 2.34 | 2.34 |
|  | Hayes-R75oth | 1.88 | 1.88 | 1.88 |

Table 6 Execution time comparison

| Algorithm | Database | Time |
|---|---|---|
| k-means | Iris | 16 |
|  | Wine | 15 |
|  | Hayes-Roth | 15 |
| Algorithm in this paper | Iris | 78 |
|  | Wine | 74 |
|  | Hayes-Roth | 56 |

In this algorithm, Euclidean distance formula is changed into kernel distance formula. In the clustering effect of spherical data, the K-means clustering algorithm performs well, but when the data points appear, the K-means clustering is poor:

The shape of the data in the two-dimensional graph is shown:

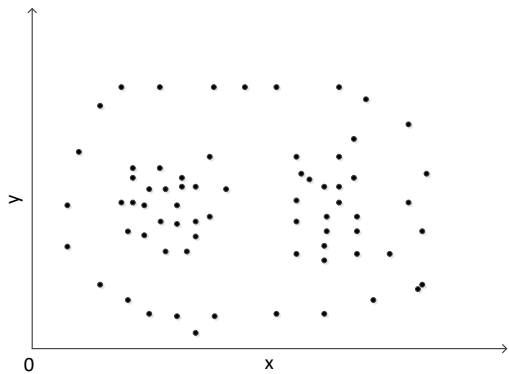

The data is divided into two kinds of effect maps under the K mean clustering algorithm, as shown in the diagram:

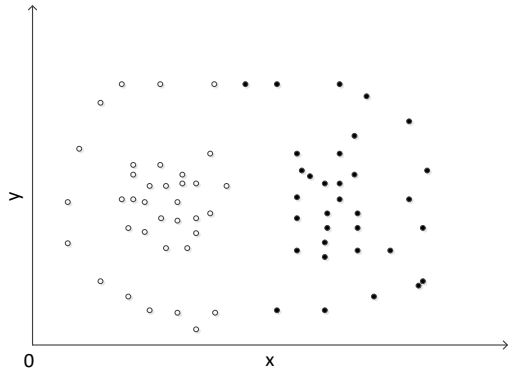

The classification effect of the data under the improved algorithm in this paper:

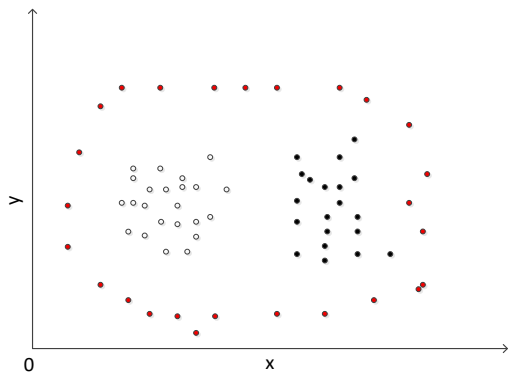

Figure 5 data classification map under the algorithm

Table 4, 5 shows that the improved algorithm for clustering accuracy of Iris and Hayes-Roth data sets have reached the clustering accuracy of traditional K-means algorithm, and the value of E is equal to the minimum value of the traditional K-means algorithm, which shows that the improved algorithm based on the density of the two sets of data with good clustering effect. For the Wine data set, the clustering accuracy is lower than the maximum accuracy of the K-means clustering algorithm, but it is higher than the minimum precision, and the E also reaches the minimum value of the traditional K-means algorithm, which shows that the clustering of Wine data sets still has good clustering results.

It can be seen from Table 6 that the improved density based K-means clustering algorithm is longer than the traditional K-means algorithm in the running time. The running time of the improved algorithm is 4.9 times, 4.9 times and 3.7 times of the running time of the traditional K-means clustering algorithm, respectively. Although this paper has some improvements on the clustering accuracy and the criterion function of the data set, but the running time is longer than the traditional algorithm, but for the case of small data sets, the execution time of the improved algorithm is still acceptable.

It can be seen from Table 4 that for the Wine data set, the accuracy of the improved algorithm does not reach the average of the traditional algorithm accuracy, but only exceeds the minimum accuracy. This is mainly because the wine data set has 14 attributes, the range of each attribute value is very large, and the distance between the objects is greatly affected, so that the k objects which are farthest from each other in the high density region can not reflect the actual distribution of the data well. It shows that the algorithm of this paper still has some limitations.

From Figure 3, 4, 5, it can be seen that, for the non spherical data, the K-means classification does not have a good classification effect. In this paper, the improved K-means algorithm, using kernel function instead of Euclidean distance, can achieve a good classification effect.

**Conclution**

    K-means clustering algorithm is a kind of traditional clustering algorithm. K-means clustering algorithm itself has some disadvantages which are too dependent on the cluster center, with the clustering center due to change of the final clustering results are not consistent; at the same time, the effect of noise data, resulting in the final clustering is not high enough and the traditional K-means clustering method it is difficult to find the globular clusters outside the cluster. The improved algorithm based on the density of K-means can eliminate the dependence on the initial cluster center, and can effectively improve the shortcomings of the traditional K-means algorithm, and has a good clustering effect.

    The computational efficiency of this algorithm is lower than the traditional K mean algorithm, and it is acceptable in the range of a certain data set.

    In this paper, the improved algorithm obtains the matlab code of the initial cluster center as follows:

1. xx=load(mdist);
2. ND=max(xx(:,2));
3. NL=max(xx(:,1));
4. if (NL>ND)
5.   ND=NL; %% To ensure that DN is the larger one in first two columns of the maximum value, and count it as the total number of data points
6. end
7. 
8. N=size(xx,1); %% xx The length of the first dimension corresponds to the number of rows in the file (i.e. the total number of distances)
9. 
10. %% Initialize to zero
11. for i=1:ND
12.   for j=1:ND
13.     dist(i,j)=0;
14.   end
15. end
16. 
17. %% xx is to assigned the dist array, pay attention to the input that only saved 0.5*DN (DN-1) values, fill it to be a full matrix
18. %% Diagonal elements are not considered here
19. for i=1:N
20.   ii=xx(i,1);
21.   jj=xx(i,2);
22.   dist(ii,jj)=xx(i,3);
23.   dist(jj,ii)=xx(i,3);
24. end
25.

26. %% Determination of dc
27.
28. percent=2.0;
29. fprintf('average percentage of neighbours (hard coded): %5.6f\n', percent);
30.
31. position=round(N*percent/100);
32. sda=sort(xx(:,3));
33. dc=sda(position);
34.
35. %% To calculate the local density *rho* (using Gaussian nuclear)
36.
37. fprintf('Computing Rho with gaussian kernel of radius: %12.6f\n', dc);
38.
39. %% Each data point rho is initialized to zero
40. for i=1:ND
41.   rho(i)=0.;
42. end
43.
44. % Gaussian kernel
45. for i=1:ND-1
46.   for j=i+1:ND
47.     rho(i)=rho(i)+exp(-(dist(i,j)/dc)*(dist(i,j)/dc));
48.     rho(j)=rho(j)+exp(-(dist(i,j)/dc)*(dist(i,j)/dc));
49.   end
50. end
51.
52. % "Cut off" kernel
53. %for i=1:ND-1
54. %  for j=i+1:ND
55. %    if (dist(i,j)<dc)
56. %      rho(i)=rho(i)+1.;
57. %      rho(j)=rho(j)+1.;
58. %    end
59. %  end
60. %end
61.
62. %% The maximum value of the matrix column is first calculated, and then the maximum value is obtained. Finally, the maximum value of all distance values is obtained
63. maxd=max(max(dist));
64.
65. %% The *rho* in descending order, *ordrho* keep order
66. [rho_sorted,ordrho]=sort(rho,'descend');
67.
68. %% The maximum value of *rho* processing data points

```matlab
69.    delta(ordrho(1))=-1.;
70.    nneigh(ordrho(1))=0;
71.
72.    %% Delta and nneigh array are generated
73.    for ii=2:ND
74.       delta(ordrho(ii))=maxd;
75.       for jj=1:ii-1
76.         if(dist(ordrho(ii),ordrho(jj))<delta(ordrho(ii)))
77.            delta(ordrho(ii))=dist(ordrho(ii),ordrho(jj));
78.            nneigh(ordrho(ii))=ordrho(jj);
79.            %% Record the number ordrho(jj) of the point which is the nearest to the ordrho(ii) in the aggregate of the data points larger than rho.
80.         end
81.       end
82.    end
83.
84.    %%   the delta value of the biggest data points of rho value is generated
85.    delta(ordrho(1))=max(delta(:));
86.
87.    %% Decision diagram
88.
89.    disp('Generated file:DECISION GRAPH')
90.    disp('column 1:Density')
91.    disp('column 2:Delta')
92.
93.    fid = fopen('DECISION_GRAPH', 'w');
94.    for i=1:ND
95.       fprintf(fid, '%6.2f %6.2f\n', rho(i),delta(i));
96.    end
97.
98.    %% Select a rectangle that surrounds the center of the class
99.    disp('Select a rectangle enclosing cluster centers')
100.
101.          %% Each computer, the root object handle only one, is the screen, its handle is always 0
102.          %% >> scrsz = get(0,'ScreenSize')
103.          %% scrsz =
104.          %%      1      1     1280      800
105.          %% 1280 and 800 is that you set the resolution of computer, scrsz (4) is 800, scrsz (3) is 1280
106.          scrsz = get(0,'ScreenSize');
107.
108.          %% For a specified location, the feeling was not so auto the: -)
109.          figure('Position',[6 72 scrsz(3)/4. scrsz(4)/1.3]);
110.
```

```
111.      %% Ind and gamma is not in use in follows
112.      for i=1:ND
113.         ind(i)=i;
114.         gamma(i)=rho(i)*delta(i);
115.      end
116.
117.      %% rho 和 delta are used to draw a "decision diagram"
118.
119.      subplot(2,1,1)
120.      tt=plot(rho(:),delta(:),'o','MarkerSize',5,'MarkerFaceColor','k','MarkerEdgeColor','k');
121.      title ('Decision Graph','FontSize',15.0)
122.      xlabel ('\rho')
123.      ylabel ('\delta')
124.
125.      subplot(2,1,1)
126.      rect = getrect(1);
127.      rhomin=rect(1);
128.      deltamin=rect(2); %%
129.
130.      %% Initialize the cluster number
131.      NCLUST=0;
132.
133.      %% cl is an array of belonging mark, cl (i) =j said the i number data points belongs to the jth cluster
134.      %% Firstly, the cl is unified initialized to -1
135.      for i=1:ND
136.         cl(i)=-1;
137.      end
138.
139.      %% The number of data points (i.e., clustering centers) is counted in the rectangular region
140.
141.      for i=1:ND
142.        if ( (rho(i)>rhomin) && (delta(i)>deltamin))
143.           NCLUST=NCLUST+1;
144.           cl(i)=NCLUST; %% The ith data point belongs to the NCLUSTth cluster
145.           icl(NCLUST)=i;%% The inverse mapping, the NCLUSTth cluster center is the ith data point.
146.        end
147.      end
148.
149.      fprintf('NUMBER OF CLUSTERS: %i \n', NCLUST);
150.
151.      disp('Performing assignment')
```